\documentclass[letterpaper]{article} % DO NOT CHANGE THIS
\usepackage{aaai25}  % DO NOT CHANGE THIS
\usepackage{times}  % DO NOT CHANGE THIS
\usepackage{helvet}  % DO NOT CHANGE THIS
\usepackage{courier}  % DO NOT CHANGE THIS
\usepackage[hyphens]{url}  % DO NOT CHANGE THIS
\usepackage{graphicx} % DO NOT CHANGE THIS
\usepackage{natbib}  % DO NOT CHANGE THIS AND DO NOT ADD ANY OPTIONS TO IT
\usepackage{caption} % DO NOT CHANGE THIS AND DO NOT ADD ANY OPTIONS TO IT
\usepackage{algorithm}
\usepackage{algorithmic}

\usepackage{amssymb}
\usepackage{amsmath}
\usepackage{makecell}
\usepackage{multirow}
\usepackage{booktabs}
\usepackage{colortbl}
\usepackage{pifont}
\definecolor{mygray}{gray}{.90}

\usepackage{newfloat}
\usepackage{listings}
\DeclareCaptionStyle{ruled}{labelfont=normalfont,labelsep=colon,strut=off} % DO NOT CHANGE THIS
\floatstyle{ruled}
\newfloat{listing}{tb}{lst}{}
\floatname{listing}{Listing}
\title{Diversifying Query: Region-Guided Transformer for 
\\ Temporal Sentence Grounding}
\author{
    Xiaolong Sun\textsuperscript{\rm 1}\equalcontrib, Liushuai Shi\textsuperscript{\rm 1}\equalcontrib, Le Wang\textsuperscript{\rm 1}\thanks{Corresponding author.}\\ Sanping Zhou\textsuperscript{\rm 1}, Kun Xia\textsuperscript{\rm 1}, Yabing Wang\textsuperscript{\rm 1}, Gang Hua\textsuperscript{\rm 2}
}
\affiliations{
    \textsuperscript{\rm 1}National Key Laboratory of Human-Machine Hybrid Augmented Intelligence,\\
    National Engineering Research Center for Visual Information and Applications,\\
    and Institute of Artificial Intelligence and Robotics, Xi’an Jiaotong University\\
    \textsuperscript{\rm 2}Multimodal Experiences Research Lab, Dolby Laboratories\\
    \{sunxiaolong,shiliushuai,xiakun\}@stu.xjtu.edu.cn, \{lewang,spzhou\}@xjtu.edu.cn, \{wyb7wyb7,ganghua\}@gmail.com
}

\usepackage{bibentry}
\begin{document}

\maketitle

\begin{abstract}
Temporal sentence grounding is a challenging task that aims to localize the moment spans relevant to a language description. Although recent DETR-based models have achieved notable progress by leveraging multiple learnable moment queries, they suffer from overlapped and redundant proposals, leading to inaccurate predictions. We attribute this limitation to the lack of task-related guidance for the learnable queries to serve a specific mode. Furthermore, the complex solution space generated by variable and open-vocabulary language descriptions complicates optimization, making it harder for learnable queries to adaptively distinguish each other, leading to more severe overlapped proposals. To address this limitation, we present the Region-Guided TRansformer (RGTR) for temporal sentence grounding, which introduces regional guidance to increase query diversity and eliminate overlapped proposals. Instead of using learnable queries, RGTR adopts a set of anchor pairs as moment queries to introduce explicit regional guidance. Each moment query takes charge of moment prediction for a specific temporal region, which reduces the optimization difficulty and ensures the diversity of the proposals. In addition, we design an IoU-aware scoring head to improve proposal quality. Extensive experiments demonstrate the effectiveness of RGTR, outperforming state-of-the-art methods on three public benchmarks and exhibiting good generalization and robustness on out-of-distribution splits. Codes are available at \url{https://github.com/TensorsSun/RGTR}
\end{abstract}

%     \link{Code}{https://github.com/TensorsSun/RGTR}
%     % \link{Datasets}{https://aaai.org/example/datasets}
%     % \link{Extended version}{https://aaai.org/example/extended-version}

\section{Introduction}
\label{intro}

Temporal sentence grounding (TSG) aims at localizing the moment spans semantically aligned with the given language description in an untrimmed video. Early methods address the TSG task by designing predefined dense proposals~\cite{gao2017tall,wang2022negative} or directly learning sentence-frame interactions~\cite{liu2022memory,yang2022entity}. The recent success of detection transformer (DETR) has inspired the integration of transformers into the TSG framework~\cite{moon2023query, xiao2024bridging}. By decoding moment spans from a set of learnable queries, they streamline the complicated grounding pipeline. % This area of research has drawn increasing attention in recent years due to its wide range of potential applications, such as human-computer interaction and information retrieval.

\begin{figure*}[!ht]
    \centering
    \includegraphics[height=5.8cm]{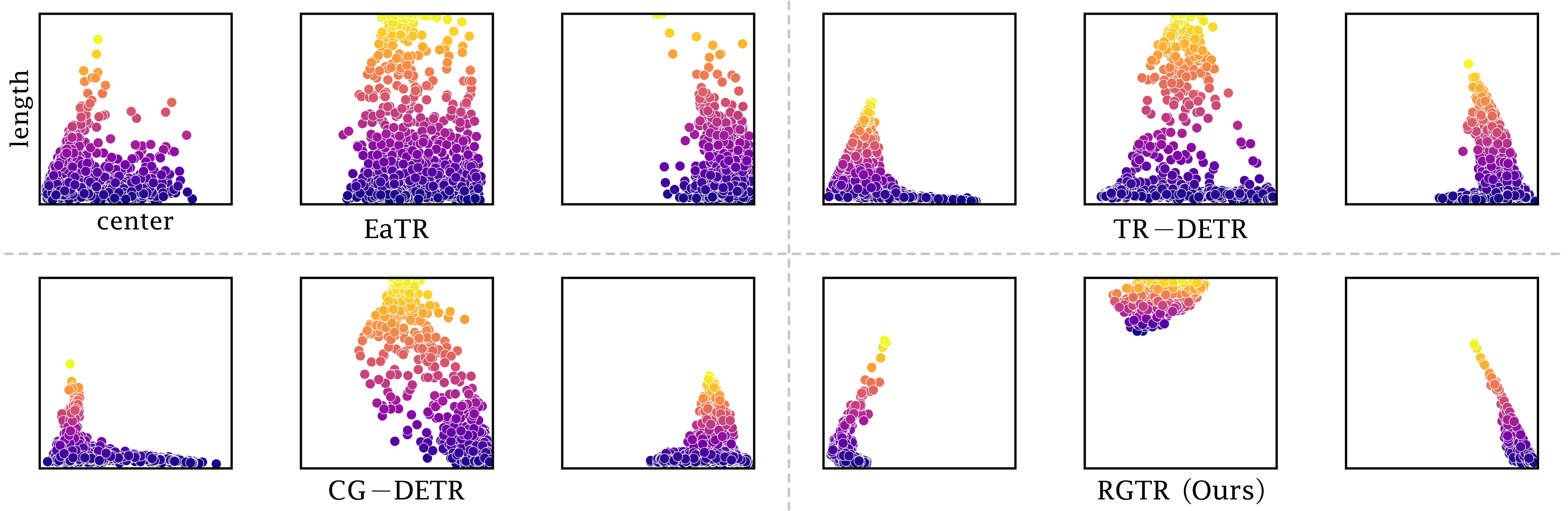}
    \caption{Visualization comparison of all moment predictions on QVHighlights \textit{val} split, for the 3 representative moment queries in EaTR~\cite{jang2023knowing}, TR-DETR~\cite{sun2024tr}, CG-DETR~\cite{moon2023correlation} and RGTR (Ours). x-axis denotes the normalized moment span center coordinate, y-axis denotes the normalized moment span length. All queries in previous methods generate numerous overlapped proposals. For example, the second query tends to predict long moments near the middle of the videos (higher middle area), but the proposals of short moments (lower area) conflict with this purpose, leading to ineffective predictions. In contrast, the predicted region of each query in our RGTR is distinct and more concentrated.}
    \label{fig:1}
    % \vspace{-0.2cm}
\end{figure*}

Although DETR-based approaches have achieved notable performance in TSG task, we still observe some unique limitations of the DETR structure compared to other fields (\textit{e.g.}, object detection). Specifically, % the proposals of different learnable queries in existing methods tend to overlap and lack diversity. 
they suffer from limited query distribution and overlapped proposals, leading to inaccurate predictions. As shown in Fig.~\ref{fig:1}, we present the center-length distribution of moment queries in three DETR-based methods, each query learns to predict different temporal regions (\textit{e.g.}, the lower left area represents a short moment near the video's start and the higher middle area represents a long moment). In previous methods, each query includes numerous overlapped and redundant proposals for the same region (\textit{e.g.}, the short moments in the lower part), resulting in ineffective predictions. % We attribute this limitation to the characteristics of the TSG task， namely the extensive variability of language descriptions and the absence of category constraints compared to other detection tasks. Moments further left (right) are closer to the video's start (end); moments higher (lower) last longer (shorter).
We attribute this limitation to the lack of task-related guidance (\textit{e.g.}, category constraints, spatial distribution prior, etc.) for the learnable queries to serve a specific mode. Although task-related guidance is crucial to reducing the overlapped proposals, it has been scarcely explored in TSG task. Furthermore, the complex solution space generated by variable and open-vocabulary language descriptions exacerbates the optimization difficulty, making it harder for learnable queries to distinguish each other adaptively and resulting in more severe overlapped proposals. Another limitation is that the proposal scoring in previous methods is purely based on the classification confidence, ignoring the quality of the predicted boundary. Instead, we argue that correctly classified proposals that better overlap with the ground-truth should be assigned higher scores. The above limitations significantly restrict the accurate localization of the DETR structure in TSG task.

In this paper, we introduce an effective Region-Guided TRansformer (RGTR) framework to cope with the aforementioned limitations in TSG task. To address the issue of overlapped proposals, %we propose a new concept of anchor pairs to replace the learnable queries in the decoder.
we introduce regional priors based on the distribution of ground-truth moment spans as task-related guidance. This regional guidance can eliminate overlapped proposals by increasing query diversity. Specifically, we design a region-guided decoder with a new concept of anchor pairs as moment queries to provide regional guidance. Each moment query consists of a static anchor and a dynamic anchor, both of which are initialized by different clustering centers on the ground-truth moment spans. Such explicit initialization imposes regional priors as guidance on each moment query, enhancing the diversity of query distribution. The two types of anchors serve different roles in the decoder. The static anchor is designed to maintain the regional guidance, so it is not updated during decoding. With the help of the fixed static anchor, the dynamic anchor continuously updates to make diverse predictions for various temporal regions. They collaboratively guide localization with explicit regional guidance and eliminate overlapped proposals. In addition, to improve the scoring of high-quality proposals, we propose an IoU-aware scoring head. By supervising the IoU score with L2 loss, the prediction head considers both classification confidence and localization quality.

Extensive experiments on three TSG benchmarks demonstrate the effectiveness of RGTR framework. As shown in Fig.~\ref{fig:1}, RGTR eliminates redundant proposals and exhibits diverse query distributions compared to previous methods.  Our main contributions are summarized as follows: (1) We design a novel region-guided decoder, which adopts a set of explicitly initialized anchor pairs as moment queries to introduce regional priors as task-related guidance. (2) We propose an IoU-aware scoring head that incorporates localization quality to enhance classification confidence estimation and distinguish high-quality proposals. (3) By employing these techniques, we introduce a Region-Guided TRansformer that eliminates overlapped proposals and improves localization quality. RGTR achieves state-of-the-art performance on three challenging benchmarks and exhibits good generalization and robustness on out-of-distribution splits.
% Interestingly, the excellent performance on out-of-distribution splits also exhibits good generalization and robustness of our method.
\section{Related Work}
\label{related}
\noindent\textbf{Temporal Sentence Grounding.}\;
Temporal sentence grounding aims at predicting the moment spans of the described activity given an untrimmed video and a language description, which is first proposed in~\cite{gao2017tall}. Early methods fall into proposal-based methods and proposal-free methods. Proposal-based methods~\cite{liu2018cross,xia2022learning,wang2022negative} initially generate multiple candidate proposals and rank them based on their similarity with the description. %Despite achieving promising performance, these methods are greatly limited by the high computational cost of proposal matching. 
Proposal-free methods~\cite{lu2019debug,chen2020rethinking,yang2022entity} are proposed to avoid the need for predefined candidate moments. Instead of relying on segment candidates, they directly predict the start and end boundaries of the target moments. %by leveraging cross-modal interactions between video and sentence. 
The recent success of detection transformer (DETR)~\cite{carion2020end} has inspired the integration of transformers into the temporal sentence grounding framework~\cite{lei2021detecting, liu2022umt, lee2023bam}. DETR-based methods simplify the whole process into an end-to-end manner by removing handcrafted techniques. % For example, EaTR~\cite{jang2023knowing} formulates an event-aware dynamic moment query to enable the model to take the input-specific content and positional information of the video into account. 
However, due to the lack of task-related guidance for the learnable queries to serve a specific mode, almost all previous methods generate numerous overlapped and redundant proposals. In contrast, our method eliminates overlapped proposals by introducing regional guidance.

\begin{figure*}[!ht]
    \centering
    \includegraphics[height=5.8cm]{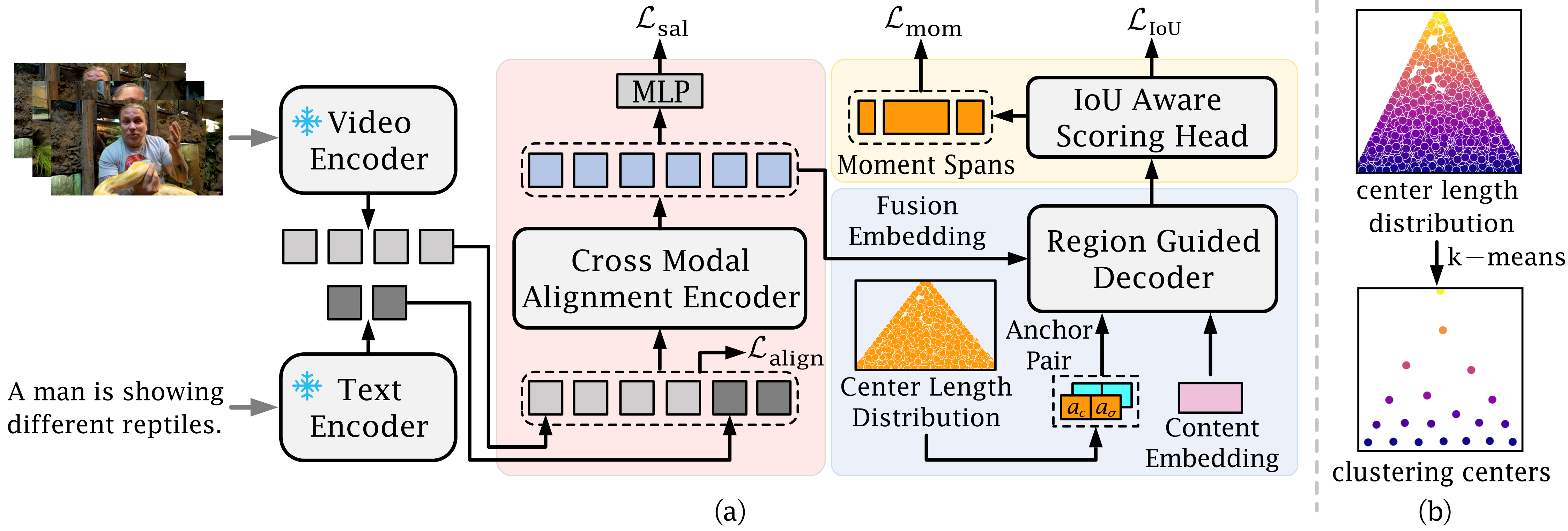}
    \caption{(a) Overview of the proposed RGTR architecture. Given a video and a text description, we first employ two frozen pre-trained models to extract visual and textual features. Subsequently, the cross-modal alignment encoder is constructed to align and fuse the visual and textual features effectively. Then, we design a region-guided decoder to introduce the regional guidance for decoding process through a set of explicitly initialized anchor pairs. Finally, the IoU-aware scoring head generates high-quality proposals by incorporating localization quality to enhance the classification confidence estimation. (b) The clustering centers with regional priors are obtained by adopting k-means algorithm on the distribution of all ground-truth moment spans.}
    \label{fig:2}
    % \vspace{-0.2cm}
\end{figure*}

\noindent\textbf{Detection Transformers.}\;
% Recently, Transformer~\cite{vaswani2017attention} has raised great attention in %natural language processing~\cite{devlin2018bert,bao2021beit} and 
% computer vision~\cite{dosovitskiy2020image,wang2018non,carion2020end}.
Recently, the adoption of transformers to object detection (DETR)~\cite{carion2020end} builds a fully end-to-end object detection system based on transformers. %which largely simplifies the traditional detection pipeline. % It also achieves notable performances compared with highly optimized CNN-based detectors. % Since the advent of DETR, a number of variants have been introduced to improve it~\cite{zhu2020deformable,meng2021conditional,zhang2022dino}.
The formulation of decoder queries has also been widely studied in previous work~\cite{zhu2020deformable,shi2022motion,shi2023trajectory}. Anchor DETR~\cite{wang2022anchor} initializes queries based on anchor points for specific detection modes. DAB-DETR~\cite{liu2022dab} formulates decoder queries with content and action embeddings. DINO~\cite{zhang2022dino} adds position priors for the positional query and randomly initializes the content query. Motivated by their great success, we introduce a set of anchor pairs %which consists of a static anchor and a dynamic anchor, 
to introduce explicit regional guidance for accurate prediction.

\section{Method}
\label{method}

% In this section, we briefly present an overview of our proposed framework in Sec.~\ref{3.1}. Then we elaborate on the main components of RGTR, including cross-modal alignment encoder (in Sec.~\ref{3.2}), region-guided decoder (in Sec.~\ref{3.3}) and IoU-aware scoring head (in Sec.~\ref{3.4}). Finally, the training objectives are introduced in Sec.~\ref{3.5}.

\subsection{Overview} \label{3.1}
Given an untrimmed video $\mathcal{V} = \left \{ {v_{t}} \right \}^{L}_{t=1}$ with $L$ frames and an associated natural language description $ \mathcal{T} = \left \{ {t_{n}} \right \}^{N}_{n=1}$ with $N$ words, TSG aims to accurately predict the moment span $m = (m_c, m_\sigma)$ that is most relevant to the given description, where $m_c$ and $m_\sigma$ represent the center time and duration length of the moment span. % Note that a video may have multiple moments that match the sentence.

Recent DETR-based methods replace hand-crafted components with learnable positional queries to predict target moments. % The query embedding, representing learnable referential search areas, consists of the content part and the positional part. Previous methods initialize content parts as zero embeddings~\cite{moon2023query,sun2024tr,yang2024task,xiao2024bridging} or dynamic event units~\cite{jang2023knowing}, and positional parts as random learnable embeddings. 
These positional queries, representing a set of learnable referential search areas, are initialized as random learnable embeddings in the previous methods~\cite{moon2023query,yang2024task,xiao2024bridging}. However, due to the lack of task-related guidance (\textit{e.g.}, categories constraints, spatial distribution prior, etc.) and the extensive variability of language descriptions, the random initialization of positional queries greatly exacerbates the optimization difficulty and produces numerous overlapped proposals.

To address this problem, we propose the Region-Guided TRansformer (RGTR), which adopts a set of explicitly initialized anchor pairs as moment queries to replace randomly initialized learnable queries without guidance. In our framework, we construct a region-guided decoder through anchor pairs to provide directive and diverse reference search areas for decoding process. In addition, we introduce an IoU-aware scoring head to distinguish high-quality proposals. The overall architecture is shown in Fig.~\ref{fig:2}a.

\subsection{Cross-Modal Alignment Encoder} \label{3.2}

Following previous methods~\cite{moon2023query,li2024momentdiff}, we use the pre-trained CLIP~\cite{radford2021learning} and Slowfast model~\cite{feichtenhofer2019slowfast} to extract clip-level visual features $F_v \in \mathbb{R}^{L\times d_v}$, where $L$ represents the number of clips and $d_v$ is the dimension of visual features. Furthermore, we utilize the CLIP model to extract word-level textual features $F_t \in \mathbb{R}^{N\times d_t}$, where $N$ is the number of words and $d_t$ is the dimension of textual features. % Note that the parameters of these models remain frozen during training.

% Following previous methods~\cite{moon2023query,sun2024tr,li2024momentdiff}, we employ frozen pre-trained models for token-level unimodal feature extraction. The pre-trained VGG model~\cite{simonyan2014very}, CLIP~\cite{radford2021learning} and Slowfast model~\cite{feichtenhofer2019slowfast} are utilized to extract clip-level visual features $F_v \in \mathbb{R}^{L\times d_v}$, where and $d_v$ is the visual feature dimension. The pre-trained Glove model~\cite{pennington2014glove} and CLIP model are utilized to extract word-level textual features $F_t \in \mathbb{R}^{N\times d_t}$, where $d_t$ is the textual feature dimension.

Given the clip-level visual features $F_v$ and the word-level textual features $F_t$, they are first projected into the common multimodal space using multi-layer perceptrons (MLPs) to produce the corresponding features $\bar{F}_v \in \mathbb{R}^{L\times D}$ and $\bar{F}_t \in \mathbb{R}^{N\times D}$, where $D$ is the embedding dimension. As highlighted in previous work~\cite{li2021align,sun2024tr}, aligning modalities before interaction could reduce the modal gap and obtain better modal representations. % Therefore, we employ an alignment loss $\mathcal{L}_{align}$ to promote the similarity of global representations of paired videos and sentences.
Therefore, we employ an alignment loss $\mathcal{L}_{\text{align}}$ to facilitate the alignment between videos and sentences.
\begin{equation}
    \mathcal{L}_{\text{align}} = -\frac{1}{B} \sum_{i=1}^{B} \log \frac{\exp((G^i_v)(G^i_t)^\top)}{\textstyle \sum_{i=1}^{B} \textstyle \sum_{j=1}^{B} \exp((G^i_v)(G^j_t)^\top)},
\end{equation}
% where $B$ is the batch size, $G^i_v \in \mathbb{R}^{D}$ and $G^i_t \in \mathbb{R}^{D}$ are the global feature of the $i$-th video and the $i$-th sentence in a training batch, respectively. We also utilize a simple visual feature refinement module~\cite{sun2024tr} to suppress the interference of text-irrelevant information in visual features.
where $B$ represents the batch size, $G^i_v \in \mathbb{R}^{D}$ and $G^i_t \in \mathbb{R}^{D}$ denote the global feature of the $i$-th video and the $i$-th sentence in a training batch, respectively. % Additionally, we employ a visual feature refinement module~\cite{sun2024tr} to suppress the interference of visual information irrelevant to the sentence.

After alignment, we adopt a text-to-video encoder to obtain text-aware video representations. Specifically, three cross-attention layers are utilized to integrate textual features into the visual features:
\begin{equation}
    \hat{F}_v = \text{Attention}(Q_v, K_t, V_t) = \text{Softmax}(\frac{Q_v K^\top_t}{\sqrt{D}})V_t.
    \label{cross_encoder}
\end{equation}
where $Q_v = \text{Linear}_q (\bar{F}_v)$, $K_t = \text{Linear}_k (\bar{F}_t)$ and $V_t = \text{Linear}_v (\bar{F}_t)$. Subsequently, three self-attention layers are leveraged to enhance the representations to help the model better understand the video sequence relations. Here, we project $\hat{F}_{v}$ to $Q_{\hat{v}}$, $K_{\hat{v}}$ and $V_{\hat{v}}$ and use them to obtain the final cross-modal fusion embedding $F$, which is imposed by saliency score constraints $\mathcal{L}_{\text{sal}}$~\cite{moon2023query}.
%Refer to the supplemental material % Sec.~\ref{A} for detailed information about $\mathcal{L}_{\text{sal}}$.

\subsection{Region-Guided Decoder} \label{3.3}
Given the fusion embedding $F$, we aim to localize moment spans semantically aligned with the description in the decoder. As discussed in Sec.~\ref{3.1}, previous methods employ randomly initialized learnable queries without task-related guidance, leading to increasing optimization difficulty and numerous overlapped proposals. In contrast, we design a region-guided decoder, which adopts a set of explicitly initialized anchor pairs as moment queries to provide directive and diverse regional guidance. Each anchor pair consists of a static anchor and a dynamic anchor, both of which are initialized by clustering centers on the ground-truth moment spans. The two types of anchors serve different roles in the decoder, where static anchors maintain regional guidance without updating and dynamic anchors make diverse predictions. They collaboratively guide localization with explicit regional guidance. The structure of the region-guided decoder is described in Fig.~\ref{fig:3}. We elaborate on the detailed process in the following. 

\noindent\textbf{Anchor Explicit Initialization.}\;
Due to the specificity of the TSG task, we lack the task-related guidance (\textit{e.g.}, category constraints) present in other detection tasks. Nonetheless, we can still provide regional guidance for the decoding process by considering the distribution of ground-truth moment spans. Specifically, the forms of static anchors and dynamic anchors are first defined as $a = (a_c, a_\sigma)$, where $a_c$ is the center coordinate and the $a_\sigma$ is the duration of the moment. Then, as shown in Fig.~\ref{fig:2}b, we generate $\mathcal{K}$ clustering centers $A\in \mathbb{R}^{\mathcal{K}\times 2}$ by adopting k-means clustering algorithm on the distribution of all ground-truth moment spans. These clustering centers represent explicit temporal regions with diverse center coordinates and durations. Since events described in the text can occur anywhere in videos, generating diverse temporal regions as guidance is crucial. Therefore, the static and dynamic anchors are initialized by $\mathcal{K}$ clustering centers: $A^0_s = A^0_d = A \in \mathbb{R}^{\mathcal{K}\times 2}$, and the positional embeddings of anchor pairs are generated by:
\begin{figure}[t]
    \centering
    \includegraphics[width=8.0cm]{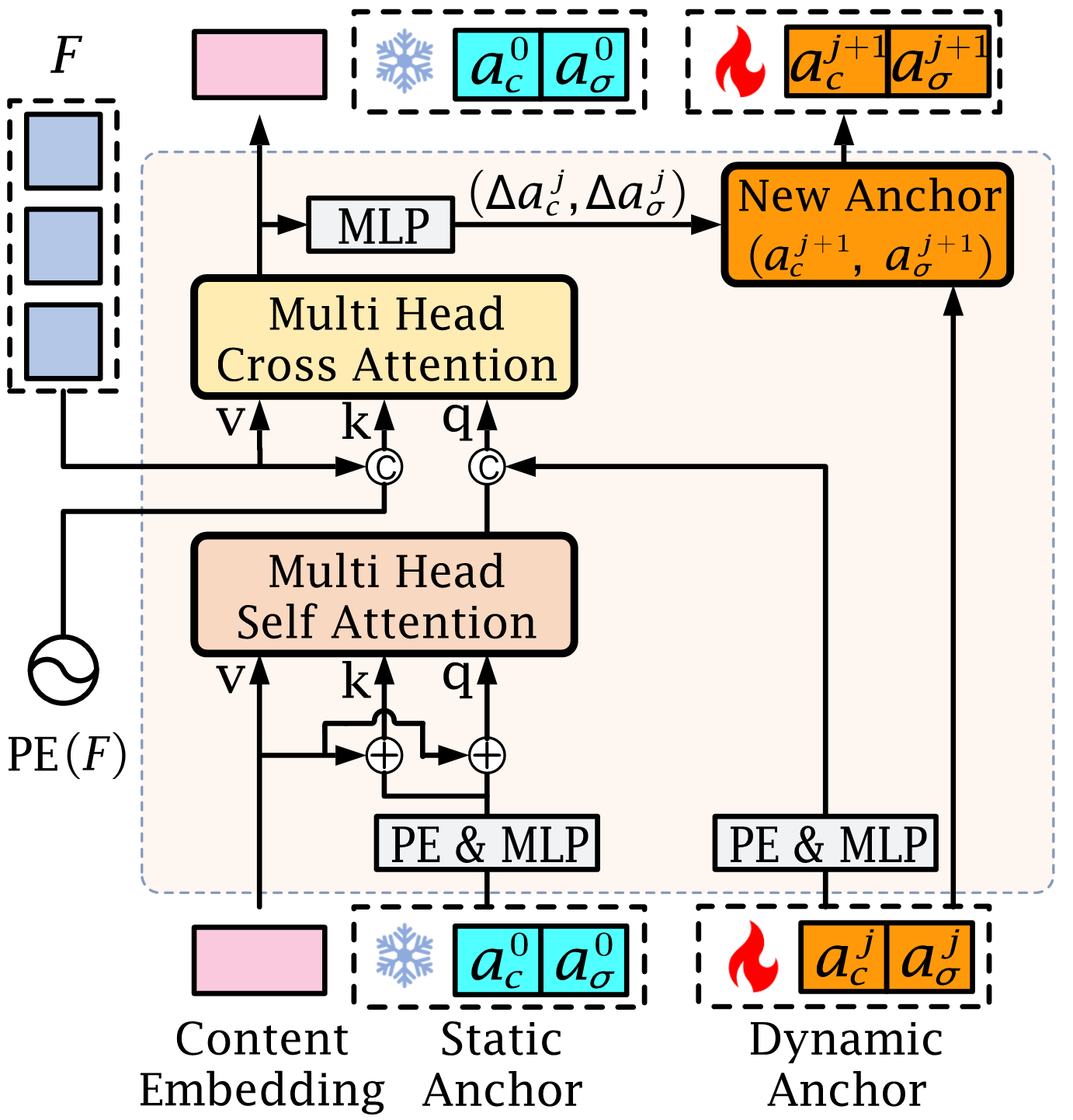}
    \caption{The structure of our proposed region-guided decoder with anchor pair (static anchor and dynamic anchor), where PE means positional encoding.}
    \label{fig:3}
    % \vspace{-0.3cm}
\end{figure}
\begin{equation}
    P^0_s = P^0_d = \text{MLP}(\text{PE}(A)),
\end{equation}
where PE($\cdot$) means positional encoding to generate sinusoidal embeddings. For clarity, we use $A^j_s$ and $P^j_s$ to sign the static anchor and its positional embedding in $j$-th decoder layer, even though it is never updated. With the explicit initialization, regional priors are introduced to guide the decoder in generating non-overlapped proposals.

\noindent\textbf{Anchor Pair Update.}\;
Although introducing regional guidance by explicit initialization, maintaining the guidance during decoding iterations is also important. Following this idea, static anchors are designed to maintain guidance without updating, while dynamic anchors are designed to update for localization as shown in Fig.~\ref{fig:3}. For static anchors,
\begin{equation}
    A^{j+1}_s = A^{0}_s = A, \;\; P^{j+1}_s = P^0_s = \text{MLP}(\text{PE}(A)).
\end{equation}
Given dynamic anchors $A^j_d = (a_c^j, a_\sigma^j)$ in $j$-th decoder layer and the relative positions $\Delta A^j_d$ = ($\Delta a^j_c, \Delta a^j_\sigma$) from a prediction head, the dynamic anchors are updated as:
\begin{equation}
\begin{split}
    A^{j+1}_d = A^j_d + \Delta &A^j_d = (a_c^j + \Delta a^j_c, \; a_\sigma^j + \Delta a^j_\sigma), \\
    P^{j+1}_d &= \text{MLP}(\text{PE}(A^{j+1}_d)).
\end{split}
\end{equation}
Note that all prediction heads share the same parameters.

\begin{table*}[ht]
\begin{center}
\renewcommand{\arraystretch}{1.0}
\resizebox{\textwidth}{!}{
\begin{tabular}{lcccccccccc}
\toprule
\multirow{3}{*}{\makecell{ \\ Method}} & \multicolumn{5}{c}{test} & \multicolumn{5}{c}{val}\\
\cmidrule(rl){2-6} \cmidrule(rl){7-11}
& \multicolumn{2}{c}{R1} & \multicolumn{3}{c}{mAP} & \multicolumn{2}{c}{R1} & \multicolumn{3}{c}{mAP}\\
\cmidrule(rl){2-3} \cmidrule(rl){4-6} \cmidrule(rl){7-8} \cmidrule(rl){9-11}
& @0.5 & @0.7 & @0.5 & @0.75 & \multicolumn{1}{c}{Avg.} & @0.5 & @0.7 & @0.5 & @0.75 & Avg.\\
\addlinespace[1pt]
% \noalign{\smallskip}
\hline
% \noalign{\smallskip}
\addlinespace[2pt]
M-DETR~\cite{lei2021detecting} & 52.89 & 33.02 & 54.82 & 29.40 & 30.73 & 53.94 & 34.84 & - & - & 32.20\\
%UMT~\cite{liu2022umt} & 56.23 & 41.18 & 53.83 & 37.01 & 36.12 & 60.26 & 44.26 & - & - & 38.59\\
% MH-DETR~\cite{xu2023mh} & 60.05 & 42.48 & 60.75 & 38.13 & 38.38 & - & - & - & - & -\\
QD-DETR~\cite{moon2023query} & 62.40 & 44.98 & 62.52 & 39.88 & 39.86 & 62.68 & 46.66 & 62.23 & 41.82 & 41.22\\
UniVTG~\cite{lin2023univtg} & 58.86 & 40.86 & 57.60 &  35.59 & 35.47 & 59.74 & - & - & - & 36.13\\
% EaTR~\cite{jang2023knowing} & - & - & - & - & - & 61.36 & 45.79 & 61.86 & 41.91 & 41.74 \\
%MomentDiff~\cite{li2024momentdiff} & 57.42 & 39.66 & 54.02 & 35.73 & 35.95 & - & - & - & - & - \\
% MESM~\cite{liu2024towards} & 62.78 & 45.20 & 62.64 & 41.45 & 40.68 & - & - & - & - & - \\
TR-DETR~\cite{sun2024tr} & 64.66 & 48.96 & 63.98 & 43.73 & 42.62 & 67.10 & 51.48 & \underline{66.27} & 46.42 & 45.09 \\
TaskWeave~\cite{yang2024task} & - & - & - & - & - & 64.26 & 50.06 & 65.39 & \underline{46.47} & 45.38 \\
UVCOM~\cite{xiao2024bridging} & 63.55 & 47.47 & 63.37 & 42.67 & 43.18 & 65.10 & 51.81 & - & - & 45.79\\
CG-DETR~\cite{moon2023correlation} & 65.43 & 48.38 & 64.51 & 42.77 & 42.86 & \underline{67.35} & \underline{52.06} & 65.57 & 45.73 & 44.93 \\
% BAM-DETR & 62.71 & 48.64 & 64.57 & 46.33 & 45.36 & 65.10 & 51.61 & 65.41 & 48.56 & 47.61\\
% $\text{R}^2$-Tuning & 68.03 & 49.35 & 69.04 & 47.56 & 46.17 & 68.71 & 52.06 & - & - & 47.59\\
% SpikeMba & 64.13 & 49.42 & - & 43.67 & 43.79 & 65.32 & 51.33 & - & 44.96 & 44.84\\
$\text{LLMEPET}^{\dagger}$~\cite{jiang2024prior} & \textbf{66.73} & \textbf{49.94} & \underline{65.76} & \underline{43.91} & \underline{44.05} & 66.58 & 51.10 & - & - & \underline{46.24}\\
\rowcolor{mygray}
% \textbf{RGTR (n = 10)} & \textbf{66.08} & \underline{48.64} & \underline{65.78} & 42.89 & 43.09 &\underline{67.61} & \underline{52.19} & \underline{67.07} & 45.86 & 45.59 \\
\rowcolor{mygray}
\textbf{RGTR (Ours)} & \underline{65.50} & \underline{49.22} & \textbf{67.12} & \textbf{45.77} & \textbf{45.53} & \textbf{67.68} & \textbf{52.90} & \textbf{67.38} & \textbf{48.00} & \textbf{46.95}\\

\toprule
\end{tabular}
}
\caption{Performance Comparison on QVHighlights \textit{test} and \textit{val} splits. $\dagger$ indicates LLM-based method.} %with the features from Slowfast and CLIP.} %We highlight the best score in each column in \textbf{bold}, and the second best score with \underline{underline}.}
\label{table1}
\end{center}
% \vspace{-0.1cm}
\end{table*}

\begin{table*}[ht]
\begin{center}
\renewcommand{\arraystretch}{1.0}
\begin{tabular}{lcccccccc}
    \toprule
    \multirow{2}{*}{Method} & \multicolumn{4}{c}{TACoS} & \multicolumn{4}{c}{Charades-STA}\\
    \cmidrule(rl){2-5} \cmidrule(rl){6-9}
    & R@0.3 & R@0.5 & R@0.7 & mIoU & R@0.3 & R@0.5 & R@0.7 & mIoU\\
    % \noalign{\smallskip}
    \hline
    % \noalign{\smallskip}
    \addlinespace[2pt]
    % 2D-TAN~\cite{zhang2020learning} & 40.01 & 27.99 & 12.92 & 27.22 & 58.76 & 46.02 & 27.50 & 41.25 \\
    M-DETR~\cite{lei2021detecting} & 37.97 & 24.67 & 11.97 & 25.49 & 65.83 & 52.07 & 30.59 & 45.54 \\
    MomentDiff~\cite{li2024momentdiff} & 44.78 & 33.68 & - & - & - & 55.57 & 32.42 & - \\
    % QD-DETR~\cite{moon2023query} & - & - & - & - & - & 57.31 & 32.55 & - \\
    UniVTG~\cite{lin2023univtg} & 51.44 & 34.97 & 17.35 & 33.60 & 70.81 & \underline{58.01} & 35.65 & 50.10 \\
    CG-DETR~\cite{moon2023correlation} & 52.23 & \underline{39.61} & 22.23 & 36.48 & 70.43 & \textbf{58.44} & \underline{36.34} & 50.13 \\
    $\text{LLMEPET}^{\dagger}$~\cite{jiang2024prior} & \underline{52.73} & - & \underline{22.78} & \underline{36.55} & \underline{70.91} & - & \textbf{36.49} & \underline{50.25} \\
    \rowcolor{mygray}
    \textbf{RGTR (Ours)} & \textbf{53.04} & \textbf{40.31} & \textbf{24.32} & \textbf{37.44} & \textbf{72.04} & 57.93 & 35.16 & \textbf{50.32}  \\
    \toprule
    \end{tabular}
\caption{Performances Comparison on TACoS and Charades-STA. $\dagger$ indicates LLM-based method.} %with the features from Slowfast and CLIP.}
\label{table2}
\end{center}
% \vspace{-0.1cm}
\end{table*}

\noindent\textbf{Region-Guided Attention Module.}\;
Similar to the general decoder, our region-guided decoder also includes two parts: self-attention module and cross-attention module. However, we employ different anchors in two modules for varying roles, as shown in Fig.~\ref{fig:3}. In the self-attention module, static anchors are utilized to focus content embeddings on preset representative temporal regions and share information across different regions. Specifically, we utilize static anchors as the positional embedding of self-attention module, such that the updated content embedding $C_{s}^j$ is as follows:
\begin{equation}
    C_{s}^j = \text{MultiHeadAttn}(C^{j-1}+P^0_s, C^{j-1}+P^0_s, C^{j-1}),
\end{equation}
where $C^{j-1} \in \mathbb{R}^{\mathcal{K}\times D}$ is the content embedding from ($j-1$)-th decoder layer, and $C^0$ is initialized to zeros. In the cross-attention module, we employ dynamic anchors as query positional embedding to aggregate region-specific features from fusion embedding $F$ with the assistance of $C_{s}^j$. Therefore, the content embedding is updated as:
\begin{equation}
    C^j = \text{MultiHeadAttn}([C^j_{s}, P_d^j],\; [F, \text{PE}(F)],\; F),
    \label{cross_decoder}
\end{equation}
where [$\cdot$,$\cdot$] means concatenation function. By adopting anchor pairs with regional guidance, the decoder reduces the optimization difficulty and eliminates overlapped proposals.
% The output of decoder is fed to the IoU-aware scoring head for predicting the moment span, confidence score, and IoU score of each proposal.

\subsection{IoU-Aware Scoring Head} \label{3.4}
The region-guided decoder improves the quality of proposals by reducing overlapped and redundant proposals, while high-quality proposals demand not only fewer duplications but also accurate boundaries. In the previous DETR-based methods~\cite{jang2023knowing,sun2024tr}, classification confidence (foreground or background) is adopted to rank all proposals. However, a single binary classification score may inadequately assess proposal quality by overlooking temporal boundary accuracy. To distinguish high-quality proposals, we introduce an IoU-aware scoring head, which considers both localization quality and classification confidence.

Specifically, the output of the decoder is fed to an FFN and a linear layer to predict the moment span and the confidence score $p_c$. Additionally, we add a linear layer to predict the expected IoU $p_{\text{IoU}}$. Instead of scoring proposals by classification confidence alone, we score them by a joint combination of confidence and IoU score, \textit{i.e.}, the product between $p_c$ and $p_{\text{IoU}}$. We supervise the IoU score with an L2 loss to the ground-truth IoU, denoted as $\hat{g}_{\text{IoU}}$,
\begin{equation}
    \mathcal{L}_{\text{IoU}} = \left | \left | \, p_{\text{IoU}} - \hat{g}_{\text{IoU}} \, \right |  \right | ^{2}.
\end{equation}
% We ablate this choice of loss in Tab.~\ref{table6} of the experiments. 
This additional IoU score can explicitly incorporate localization quality to enhance classification confidence estimation, thereby generating high-quality proposals. Additionally, non maximum suppression (NMS) is applied during inference.

\subsection{Training Objectives} \label{3.5}
The objective losses of RGTR include four parts: moment loss $\mathcal{L}_{\text{mom}}$, saliency loss $\mathcal{L}_{\text{sal}}$, alignment loss $\mathcal{L}_{\text{align}}$ and IoU loss $\mathcal{L}_{\text{IoU}}$. % Following~\cite{moon2023query}, moment loss includes L1, gIoU, and focal loss, and saliency loss includes margin ranking loss and contrastive loss. 
The overall objective is defined as:
\begin{equation}
    \mathcal{L}_{\text{overall}} = \mathcal{L}_{\text{mom}} + \lambda_{\text{sal}}\mathcal{L}_{\text{sal}} + \lambda_{\text{align}}\mathcal{L}_{\text{align}} + \lambda_{\text{IoU}}\mathcal{L}_{\text{IoU}},
\end{equation}
where $\lambda_\ast$ are the balancing parameters. $\mathcal{L}_{\text{mom}}$ and $\mathcal{L}_{\text{sal}}$ are consistent with QD-DETR~\cite{moon2023query}.
%Refer to the supplemental material %Sec.~\ref{A} for detailed information about the training objectives.

\section{Experiments}
\label{experiments}
\subsection{Datasets and Metrics}
\label{4.1}
\noindent\textbf{Datasets.}\;
We evaluate the proposed method on three temporal sentence grounding benchmarks, including the QVHighlights~\cite{lei2021detecting}, Charades-STA~\cite{gao2017tall}, and TACoS~\cite{regneri2013grounding}. QVHighlights spans various themes, Charades-STA comprises intricate daily human activities, and TACoS mainly showcases long-form videos focusing on culinary activities.

\noindent\textbf{Metrics.}\;
We adopt the Recall@1 (R1) under the IoU thresholds of 0.3, 0.5, and 0.7. Since QVHighlights contains multiple ground-truth moments per sentence, we also report the mean average precision (mAP) with IoU thresholds of 0.5, 0.75, and the average mAP over a set of IoU thresholds [0.5: 0.05: 0.95]. For Charades-STA and TACoS, we compute the mean IoU of top-1 predictions.

\subsection{Implementation Details}
\label{4.2}
Following previous methods~\cite{moon2023query}, we use SlowFast and CLIP to extract visual features and CLIP to extract textual features. %In the encoder, we also use a local regular loss following~\cite{sun2024tr}.
We set the embedding dimension $D$ to 256. The number of anchor pairs $\mathcal{K}$ is set to 20 for QVHighlights, 10 for Charades-STA and TACoS. The NMS threshold is set to 0.8. The balancing parameters are set as: $\lambda_{\text{align}} = 0.3$, $\lambda_{\text{iou}} = 1$, and $\lambda_{\text{sal}}$ is set as 1 for QVHighlights, 4 for Charades-STA and TACoS. We train all models with batch size 32 for 200 epochs using the AdamW optimizer with weight decay 1e-4. The learning rate is set to 1e-4. %For more details please refer to supplemental material Sec.~\ref{B2}.%All the experiments are implemented on a single Nvidia RTX 3090 GPU.  and the number of attention heads to 8. The VGG~\cite{simonyan2014very} and Glove~\cite{pennington2014glove} features are also employed on out-of-distribution splits.

\setlength{\tabcolsep}{4pt}
\begin{table}[t]
\begin{center}
\begin{tabular}{lccc}
\toprule
Method & R0.5 & R0.7 & $\text{mAP}_{avg}$\\
% \noalign{\smallskip}
\hline
% \noalign{\smallskip}
\addlinespace[3pt]
\hspace{-1mm}\textit{Charades-STA-Len} \\
2D-TAN~\cite{zhang2020learning} & 28.68 & 17.72 & 22.79 \\
MMN~\cite{wang2022negative} & 34.31 & 19.94 & 26.85 \\
$\text{QD-DETR}^\dagger$~\cite{moon2023query} & \underline{54.06} & \underline{32.53} & \underline{36.37} \\
MomentDiff~\cite{li2024momentdiff} & 38.32 & 23.38 & 28.19 \\
\rowcolor{mygray}
\textbf{RGTR} &  \textbf{61.17} & \textbf{40.23} & \textbf{44.30} \\
\hline
\addlinespace[3pt]
\hspace{-1mm}\textit{Charades-STA-Mom} \\
2D-TAN~\cite{zhang2020learning} & 20.44 & 10.84 & 17.23 \\
MMN~\cite{wang2022negative} & 27.20 & 14.12 & 19.18 \\
$\text{QD-DETR}^\dagger$~\cite{moon2023query} & \underline{46.31} & \underline{28.65} & \underline{30.46} \\
MomentDiff~\cite{li2024momentdiff}& 33.59 & 15.71 & 21.37 \\
\rowcolor{mygray}
\textbf{RGTR} & \textbf{49.81} & \textbf{29.77} & \textbf{33.19} \\
\addlinespace[1pt]
\toprule
\end{tabular}
\caption{Results on two out-of-distribution splits of Charades-STA. The VGG and Glove features are employed for all models. $\dagger$ indicates reproduced by official codebase.}
\label{table3}
\end{center}
% \vspace{-0.1cm}
\end{table}

\begin{table}[t]
\begin{center}
\renewcommand{\arraystretch}{1.0}
    \begin{tabular}{ccccccc}
    \toprule
    Setting & AEI & RGAM & IASH & R0.5 & R0.7 & $\text{mAP}_{avg}$\; \\
    % \noalign{\smallskip}
    \hline
    % \noalign{\smallskip}
    \addlinespace[2pt]
    (a) & & & & 65.35 & 48.97 & 43.12  \\
    (b) & \ding{51} &  &  & 64.65 & 50.58 & 44.82 \\
    (c) & & & \ding{51} & 66.19 & 49.61 & 44.03 \\
    (d) & \ding{51} & \ding{51} &  & 65.55 & 51.29 & 45.36 \\
    (e) & \ding{51} &  & \ding{51} & 66.13 & 51.68 & 46.51 \\
    \rowcolor{mygray}
    (f) & \ding{51} & \ding{51} & \ding{51} & \textbf{67.68} & \textbf{52.90} & \textbf{46.95} \\
    \addlinespace[1pt]
    \toprule
    \end{tabular}
\caption{Ablation study on the components of RGTR on QVHighlights \textit{val} split. It investigates the anchor explicit initialization (AEI), the region-guided attention module (RGAM), and the IoU-aware scoring head (IASH).}
\label{table4}
\end{center}
% \vspace{-0.4cm}
\end{table}

\subsection{Performance Comparison}\label{4.3}
As shown in Tab.~\ref{table1}, we compare RGTR to previous methods on QVHighlights. For a fair comparison, we report numbers for both the test and validation splits. Our method achieves new state-of-the-art performance on almost all metrics. Specifically, RGTR outperforms the latest methods like LLMEPET~\cite{jiang2024prior}, achieving 67.12\% at mAP@0.5 and 45.53\% at $\text{mAP}_{avg}$ on the test split. %Particularly, the average mAP score of 45.53\% on the test dataset marks a significant improvement over UVCOM by 2.35\%. 
On the validation split, RGTR also maintains its lead. The notable performance advantages of RGTR demonstrate the effectiveness of anchor pairs with explicit regional guidance.

\begin{table}
\begin{center}
\renewcommand{\arraystretch}{1.0}
    \begin{tabular}{cccccc}
    \toprule
    & Method & Changes & R0.5 & R0.7 & $\text{mAP}_{avg}$\; \\
    % \noalign{\smallskip}
    \hline
    % \noalign{\smallskip}
    \addlinespace[2pt]
    & \multirow{3}{*}{\makecell{Initialization \\ Method}} & random & 66.19 & 49.61 & 44.03 \\
    & & uniform grid & 67.10 & 50.97 & 44.93 \\
    & & \cellcolor{mygray}k-means & \cellcolor{mygray}\textbf{67.68} & \cellcolor{mygray}\textbf{52.90} & \cellcolor{mygray}\textbf{46.95} \\
    \hline
    \addlinespace[2pt]
    & \multirow{3}{*}{\makecell{Scoring \\ Method}} & IoU superv. & \textbf{67.87} & 52.84 & 46.54 \\
    & & cls + IoU & 67.23 & 52.39 & 46.92 \\
    & & \cellcolor{mygray}cls × IoU & \cellcolor{mygray}67.68 & \cellcolor{mygray}\textbf{52.90} & \cellcolor{mygray}\textbf{46.95} \\
    \addlinespace[1pt]
    \toprule
    \end{tabular}
\caption{Ablation study on initialization and scoring method.}
\label{table5}
\end{center}
% \vspace{-0.3cm}
\end{table}

\begin{figure}[t]
    \centering
    \includegraphics[height=3.2cm]{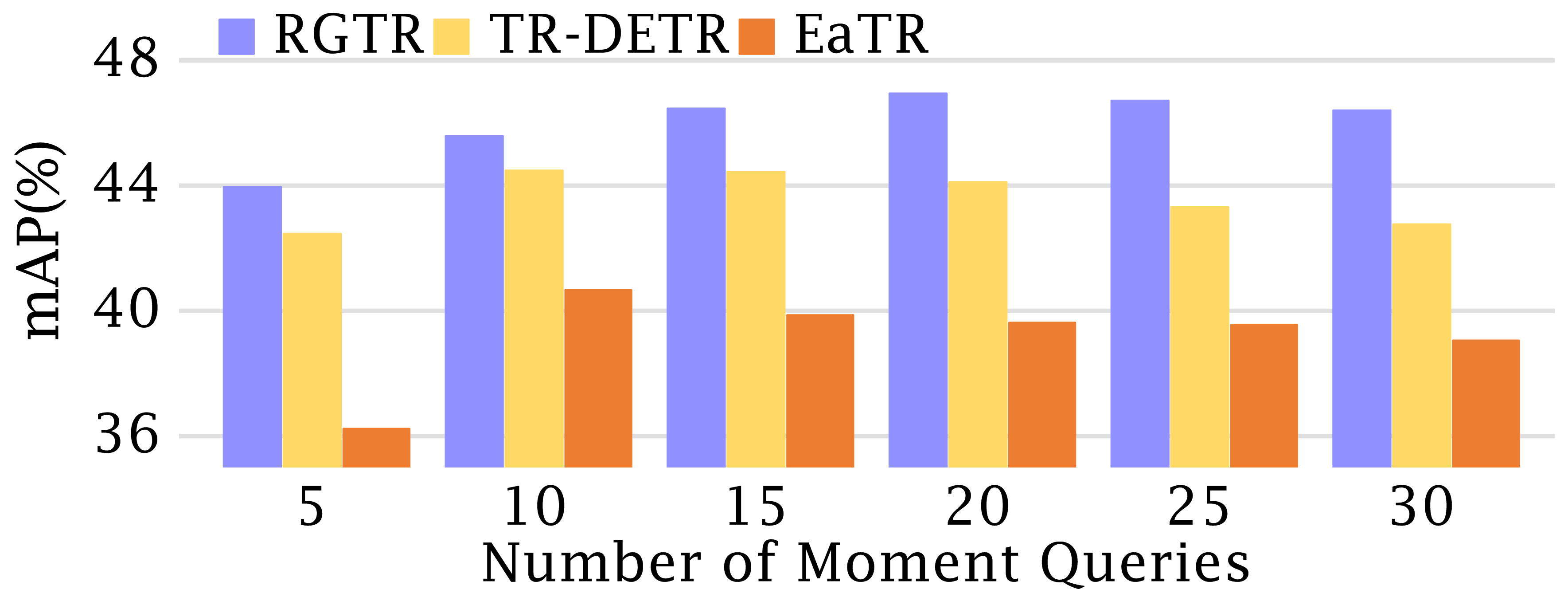}
    \caption{Ablation study on number of moment queries $\mathcal{K}$.}
    % \vspace{-0.2cm}
    \label{fig:4}
\end{figure}

\begin{figure}[!t]
    \centering
    \includegraphics[height=3.1cm]{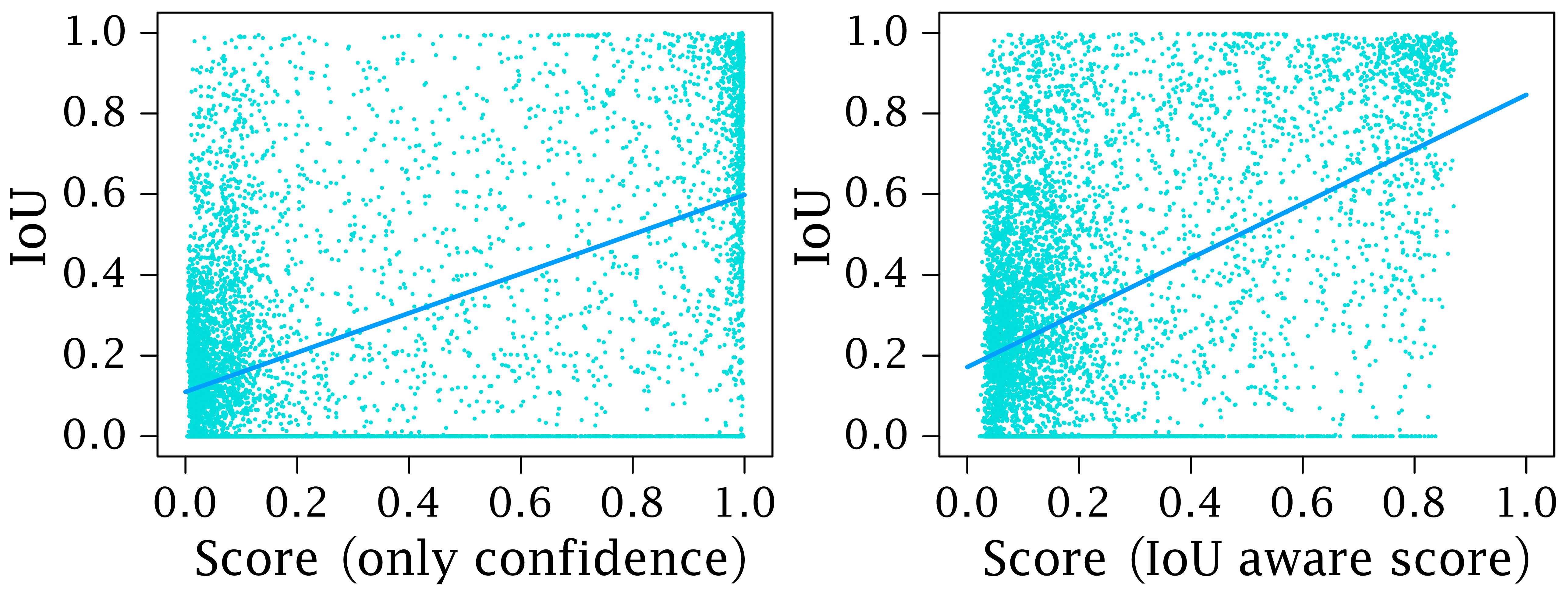}
    \caption{Correlation between scores and ground-truth IoUs.}
    % \vspace{-0.2cm}
    \label{fig:5}
\end{figure}

Tab.~\ref{table2} presents comparisons on TACoS and Charades-STA. Our method achieves the best performance on TACoS. On Charades-STA, RGTR also maintains its competitiveness. However, we observe that while our results are notably superior on QVHighlights, the margin is slightly reduced on TACoS and Charades-STA. We attribute this to the biased distribution of the two datasets compared to QVHighlights, resulting in less query diversity learned by anchor pairs. %We also provide results of RGTR on the anti-biased Charades-STA in the supplemental material.%Sec.~\ref{C1}.

\begin{figure*}[t]
    \centering
    \includegraphics[height=3.8cm]{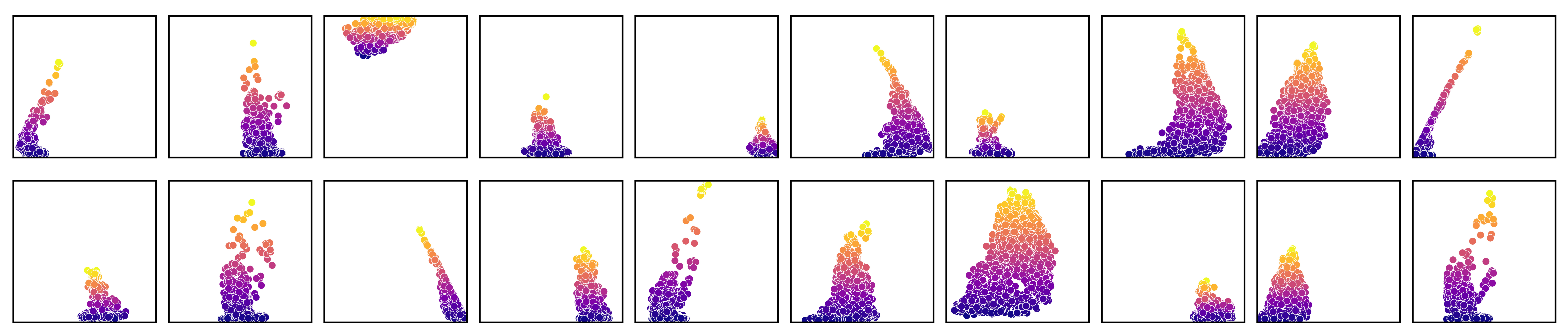}
    \caption{Visualization of moment predictions on QVHighlights \textit{val} split, for all 20 dynamic anchors in region-guided decoder.}
    \label{fig:6}
    % \vspace{-0.3cm}
\end{figure*}

\begin{figure}[t]
    \centering
    \includegraphics[height=4cm]{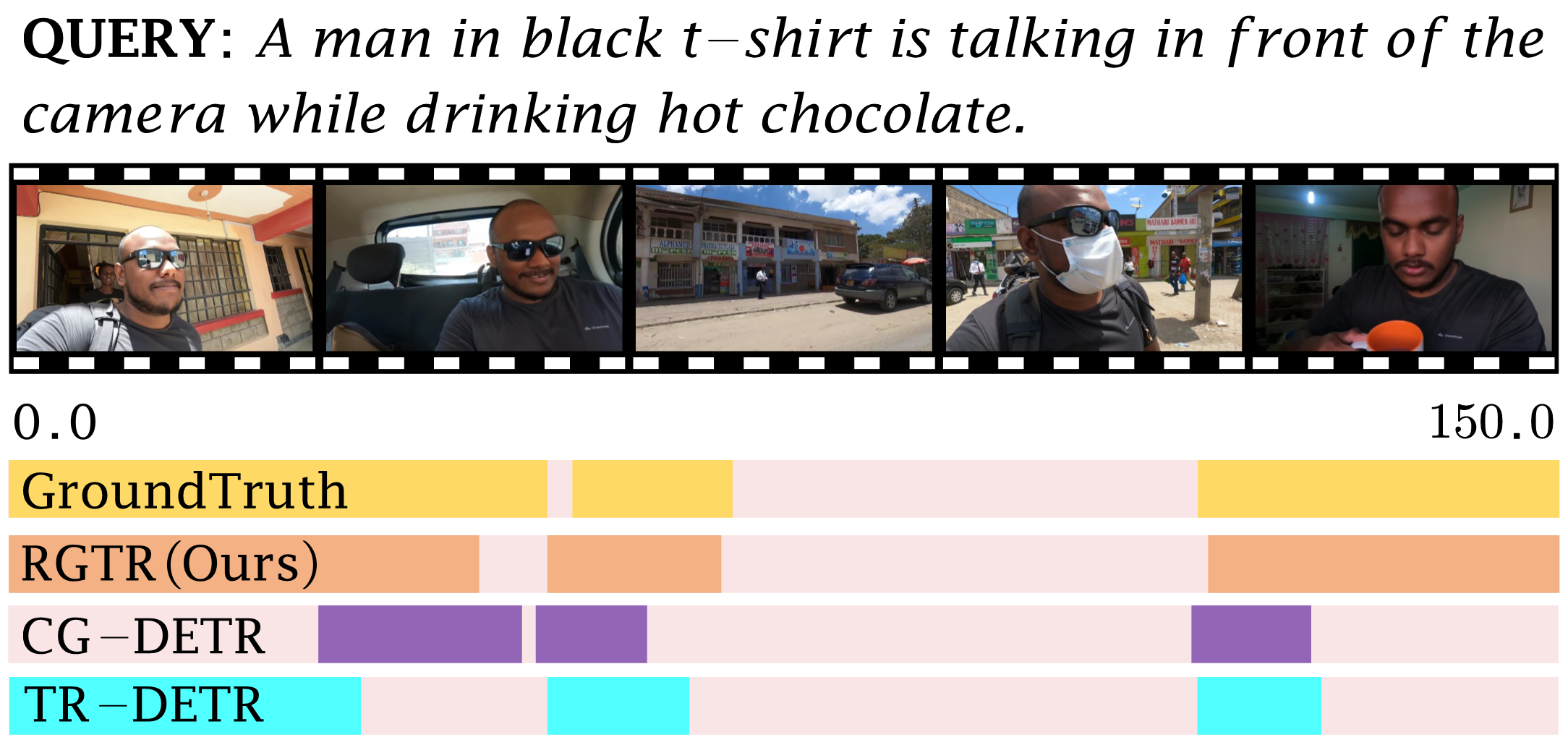}
    \caption{The qualitative result on QVHighlights.}
    \label{fig:7}
    % \vspace{-0.5cm}
\end{figure}

\subsection{Experiments on Out-of-Distribution Splits}
To measure robustness, we also evaluate RGTR on two out-of-distribution splits~\cite{li2024momentdiff}, Charades-STA-Len and Charades-STA-Mom, with distribution shifts of the length and moment location between training and test sets respectively. Since our anchor pairs are initialized by clustering centers on the training set, performance may degrade when the distribution changes significantly. However, as shown in Tab.~\ref{table3}, RGTR outperforms all previous methods under both out-of-distribution settings. Such surprising results indicate that the regional guidance introduced by anchor pairs works more by increasing the diversity between moment queries, rather than merely relying on the similarity between the training and test set distributions. Ablation experiments on other anchor initialization methods in Tab.~\ref{table5} also confirm this point. Even with uniform grid points for initialization, which also serve as an initialization method to increase query diversity but are unrelated to the dataset distribution, the model performance improves significantly. Therefore, despite a distribution shift, the query diversity from regional guidance remains crucial for effective localization. 

\subsection{Ablation Study}
%To investigate the impact corresponding to key components of the proposed method, we conduct ablation studies on the validation set of QVHighlights.

\noindent\textbf{Main Ablation.}\;
We first investigate the effectiveness of each component in RGTR. As shown in Tab.~\ref{table4}, we report the impact according to anchor explicit initialization, region-guided attention module, and IoU-aware scoring head. Notably, setting (b) represents the decoder only uses the explicitly initialized dynamic anchors, while setting (d) utilizes both static and dynamic anchors during the decoding process. The results demonstrate that each component contributes significantly to overall performance and setting (f) improves performance by 3.93\% in terms of R1@0.7 and 3.83\% in terms of mAP$_{\text{avg}}$ by using all components.

\noindent\textbf{Anchor Initialization Method.}\;
We adopt another two simple initialization methods to replace the k-means algorithm. ``random'' means utilizing random learnable queries as moment queries. ``uniform grid'' means generating a uniform grid on the normalized $m_c\times m_\sigma$ area, and uniformly sampling 5 $\times$ 5 = 25 points in a practical temporal region. As shown in Tab.~\ref{table5}, the performance of k-means initialization is significantly better than random initialization and uniform grid initialization. It verifies that k-means algorithm can provide optimal explicit regional priors for decoding process.

%\noindent\textbf{IoU loss type.}\;
%As shown in Tab.~\ref{table6}, we employ different IoU loss to supervise the IoU score. All loss types can significantly improve performance, among which L2 loss achieves optimal performance.

\noindent\textbf{Scoring Method.}\;
Tab.~\ref{table5} compares the product fusion with other scoring methods, where IoU superv. means only using confidence score with IoU loss as supervision. All methods have significant performance improvements, among which the product method achieves the best performance.

\noindent\textbf{Number of Moment Queries.}\;
In previous methods, the number of moment queries $\mathcal{K}$ is typically limited to 10. This is because increasing $\mathcal{K}$ without explicit guidance only produces more overlapped proposals, resulting in negligible performance improvement or even degradation. In contrast, our method provides explicit regional guidance for each moment query, \textit{i.e.}, each moment query is accountable for a specific temporal region. Therefore, increasing $\mathcal{K}$ allows moment queries to cover more temporal regions, leading to effective prediction. As shown in Fig.~\ref{fig:4}, we present the performance of EaTR, TR-DETR, and our RGTR in terms of mAP$_\text{avg}$ according to $\mathcal{K}$. We re-implement the other two methods in different $\mathcal{K}$. As discussed above, for TR-DETR and EaTR, performance peaks when $\mathcal{K}$ reaches 10 and then declines significantly. In contrast, for RGTR, increasing $\mathcal{K}$ to 20 significantly improves performance, demonstrating the effectiveness of anchor pairs with explicit regional guidance.

\noindent\textbf{Correlation between Score and IoU.}\;
To compare IoU-aware scoring and classification confidence scoring, we draw scatter plots of the correlation between scores and ground-truth IoUs on the QVHighlights validation set in Fig.~\ref{fig:5}. It can be observed that our IoU-aware score shows a stronger correlation with the ground-truth IoU, \textit{i.e.}, the slope of the fitted line increases from 0.49 to 0.67, improving the distinction of high-quality proposals.

\subsection{Visualization and Qualitative Result}
As shown in Fig.~\ref{fig:6}, we visualize all 20 dynamic anchors in the region-guided decoder on QVHighlights. Compared with previous methods in Fig.~\ref{fig:1}, RGTR introduces regional guidance through anchor pairs, effectively enhancing query diversity and eliminating numerous overlapped proposals.

In Fig.~\ref{fig:7}, we illustrate a qualitative example on QVHighlights, where the sentence corresponds to multiple moment spans. Since our method emphasizes enhancing query diversity, RGTR generates more accurate predictions than other methods, especially in the case of requiring simultaneous attention to different center coordinates and durations.

\section{Conclusion}
In this paper, we propose a Region-Guided TRansformer (RGTR) framework to address the limitations of DETR structure in TSG task. To eliminate overlapped proposals, we design a region-guided decoder, which adopts a set of anchor pairs as moment queries to introduce explicit regional guidance for decoding process. Each anchor pair takes charge of moment prediction for a specific temporal region, which reduces optimization difficulty and eliminates redundant proposals. To distinguish high-quality proposals, we employ an IoU-aware scoring head that incorporates localization quality to enhance classification confidence estimation. Experiments on three public datasets and two out-of-distribution splits demonstrate the superiority of RGTR.
\setcounter{secnumdepth}{0}
\section{Acknowledgments}
This work was supported in part by National Science and Technology Major Project under Grant 2023ZD0121300, National Natural Science Foundation of China under Grants 62088102, 12326608 and 62106192, Natural Science Foundation of Shaanxi Province under Grant 2022JC-41, and Fundamental Research Funds for the Central Universities under Grant XTR042021005.

\bibliography{aaai25}
\setcounter{secnumdepth}{2}
\section{Appendix}

\subsection{Details about Training Objectives}
\label{A}

As mentioned in Method, in addition to the alignment loss $\mathcal{L}_{\text{align}}$ and the IoU loss $\mathcal{L}_{\text{IoU}}$, we also adopt the moment loss $\mathcal{L}_{\text{mom}}$ and the saliency loss $\mathcal{L}_{\text{sal}}$ as supervision. To predict the timestamp of the target moments, we utilize L1 loss and generalized IoU loss with focal loss~\cite{lin2017focal} to classify the moment queries between foreground and background. Given the ground truth moment $\hat{m} = (\hat{m}_c, \hat{m}_\sigma)$ and binary classification label $\hat{g}_{\text{c}}$, and corresponding predictions as $m = (m_c, m_\sigma)$ and  $p_\text{c}$, respectively, then the moment loss $\mathcal{L}_{\text{mom}}$ is formulated as:
\begin{equation}
\begin{split}
    \mathcal{L}_{\text{mom}} = &\lambda_{\text{L1}}\left | \left | \, \hat{m} - m \, \right |  \right | \\
    &+ \lambda_{\text{gIoU}}\mathcal{L}_{\text{gIoU}}(\hat{m}, m) + \lambda_{\text{cls}}\mathcal{L}_{\text{cls}}(\hat{g}_{\text{c}}, p_\text{c}),
\end{split}
\end{equation}
where $\lambda_\ast$ are the balancing parameters. The $\mathcal{L}_{\text{cls}}$ is as follow:
\begin{equation}
\mathcal{L}_{\text{cls}}(g_{\text{c}}, p_\text{c}) = 
\begin{cases}
    -\alpha(1-p_\text{c})^{\gamma}\log{p_\text{c}}& \text{if}\;g_{\text{c}} = 1 \\
    -(1 - \alpha)p_\text{c}^{\gamma}\log{(1 - p_\text{c})}& \text{otherwise}, \\
\end{cases}
\end{equation}
where $\alpha = 0.25$ and $\gamma = 2$ are empirical hyperparameters.

Follow the previous methods~\cite{lei2021detecting,moon2023query}, we adopt the saliency loss $\mathcal{L}_{\text{sal}}$ for effective multimodal alignment in the encoder. The saliency loss consists of margin ranking loss $\mathcal{L}_{\text{margin}}$ and rank contrastive loss $\mathcal{L}_{\text{cont}}$. The margin ranking loss aims to encourage the model to produce higher saliency scores for the clips relevant to the given sentence compared to less related clips, and can be formulated as:

\begin{equation}
    \mathcal{L}_{\text{margin}} = \text{max}(0, \Delta + S(x^{\text{low}}) - S(x^{\text{high}})),
\end{equation}
where $\Delta$ is the margin, $S(\cdot)$ is the saliency score estimator, and $x^{\text{high}}$ and $x^{\text{low}}$ are video tokens from two pairs of high and low-rank clips, respectively. The rank contrastive loss is utilized to preserve the ground-truth clip ranking in prediction scores, and can be formulated as:
\begin{equation}
    \mathcal{L}_{\text{cont}} = - \sum_{r=1}^{R} \log \frac{\textstyle \sum_{x\in X_r^{\text{pos}}}\exp(S(x)/\tau)}{\textstyle \sum_{x\in (X_r^{\text{pos}}\cup X_r^{\text{neg}})} \exp(S(x)/\tau)},
\end{equation}
where $\tau$ is a temperature scaling parameter, $R$ is the maximum rank value, and $X_r^{\text{pos}}$ and $X_r^{\text{neg}}$ are the positive set and negative set built by saliency scores. Following~\cite{sun2024tr}, we inject saliency scores into the temporal grounding pipeline and use the predicted moments to refine the initial saliency distribution. Therefore, the saliency loss $\mathcal{L}_{\text{sal}}$ and the total loss can be formulated as:
\begin{equation}
    \mathcal{L}_{\text{sal}} = \mathcal{L}_{\text{margin}} + \mathcal{L}_{\text{cont}},
\end{equation}
\begin{equation}
    \mathcal{L}_{\text{overall}} = \mathcal{L}_{\text{mom}} + \lambda_{\text{sal}}\mathcal{L}_{\text{sal}} + \lambda_{\text{align}}\mathcal{L}_{\text{align}} + \lambda_{\text{IoU}}\mathcal{L}_{\text{IoU}}.
\end{equation}

\subsection{Additional Details on Experiment Settings}
\label{B}

\subsubsection{Datasets}
\textbf{QVHighlights} is a relatively recently publicized dataset by~\cite{lei2021detecting}. Consisting of varying lengths of moments and diverse text queries, it is a challenging and only dataset for joint moment retrieval and highlight detection tasks. It contains 10,148 videos and each 150 seconds long. On average, there are approximately 1.8 non-overlapping moments per query, annotated on 2s non-overlapping clips. Providing 10,310 queries with 18,367 annotated moments, it provides a test server on Codalab to ensure fair comparisons. The training set, validation set and test set include 7,218, 1,550 and 1,542 video-text pairs, respectively. ~\textbf{Charades-STA} is annotated by~\cite{gao2017tall} on Charades datasets, originally collected for video action recognition and video captioning, using semi-automatic methods. In total, the video length is 30 seconds on average. Each video is annotated with an average of 2.4 moments and the target moment length is around 8 seconds. There are 12,408 and 3,720 query-moment pairs in the training and testing sets, respectively. ~\textbf{TACoS} is collected by~\cite{regneri2013grounding} and consists of 127 videos on cooking activities, which are around 5 minutes on average. TACoS is a more challenging dataset due to the long duration of each video and the lack of scene diversity. We adopt the same split as~\cite{moon2023correlation}, which involves 9,790 pairs for training and 4,436 pairs for testing.

\begin{table}[t]
\begin{center}
\renewcommand{\arraystretch}{1.0}
\begin{tabular}{lcccc}
    \toprule
    \multirow{3}{*}{Method} & \multicolumn{2}{c}{test} & \multicolumn{2}{c}{val}\\
    \cmidrule(rl){2-3} \cmidrule(rl){4-5}
    & \multicolumn{2}{c}{$\ge$ Very Good} & \multicolumn{2}{c}{$\ge$ Very Good}\\
    \cmidrule(rl){2-3} \cmidrule(rl){4-5}
    & mAP & HIT@1 & mAP & HIT@1\\
    % \noalign{\smallskip}
    \hline
    % \noalign{\smallskip}
    \addlinespace[2pt]
    QD-DETR & 38.94 & 62.40 & 39.13 & 63.03 \\
    UniVTG & 38.20 & 60.96 & 38.83 & 61.81 \\
    TR-DETR & 39.91 & 63.42 & 40.55 & 64.77 \\
    TaskWeave & - & - & 39.28 & 63.68 \\
    UVCOM & 39.74 & 64.20 & 40.03 & 63.29 \\
    CG-DETR & 40.33 & \textbf{66.21} & \underline{40.79} & \textbf{66.71} \\
    LLMEPET & \textbf{40.33} & \underline{65.69} & - & - \\
    \rowcolor{mygray}
    \textbf{RGTR (Ours)} & 39.98 & 64.01 & \textbf{41.15} & \underline{66.13} \\
    \addlinespace[1pt]
    \toprule
    \end{tabular}
\caption{Highlight detection results on QVHighlights.}
\label{table6}
% \vspace{-0.2cm}
\end{center}
\end{table}

\begin{table}[t]
\begin{center}
\renewcommand{\arraystretch}{1.0}
\begin{tabular}{cccc}
    \toprule
    Method &  R1@0.5 & R1@0.7 & mAP$_{avg}$ \\
    % \noalign{\smallskip}
    \hline
    % \noalign{\smallskip}
    \addlinespace[2pt]
     2D-TAN & 28.95 & 12.78 & 12.60  \\
     MMN & 34.56 & 15.84 & 15.73  \\
     MomentDETR & 41.18 & 19.31 & 18.95  \\
     MomentDiff & \underline{47.17} & \underline{22.98} & \underline{22.76}  \\
    \rowcolor{mygray}
     RGTR(Ours) & \textbf{48.53} & \textbf{24.27} & \textbf{28.86} \\
    \addlinespace[1pt]
    \toprule
\end{tabular}
\caption{Results on the out-of-distribution test split of Charades-CD. The VGG and Glove features are employed.}
\label{table7}
\vspace{-0.2cm}
\end{center}
\end{table}

\begin{table}[t]
\begin{center}
\renewcommand{\arraystretch}{1.0}
\begin{tabular}{lcc}
    \toprule
    Method & R1@0.5 & R1@0.7\\
    % \noalign{\smallskip}
    \hline
    % \noalign{\smallskip}
    \addlinespace[2pt]
    2D-TAN & 40.94 & 22.85 \\
    FVMR & 42.36 & 24.14 \\
    SSRN & 46.72 & 27.98 \\
    UMT & 48.31 & 29.25 \\
    MomentDiff & 51.94 & 28.25 \\
    QD-DETR & 52.77 & 31.13 \\
    TR-DETR & 53.47 & 30.81 \\
    % TaskWeave~\cite{yang2024task} & 56.51 & 33.66 \\
    CG-DETR & \underline{55.22} & \underline{34.19} \\
    \rowcolor{mygray}
    \textbf{RGTR (Ours)} & \textbf{55.48} & \textbf{34.33} \\
    \addlinespace[1pt]
    \toprule
\end{tabular}
\caption{Results on Charades-STA with VGG backbone.}
\label{table8}
% \vspace{-0.2cm}
\end{center}
\end{table}

\begin{table}[t]
\begin{center}
\renewcommand{\arraystretch}{1.0}
\begin{tabular}{cccc}
    \toprule
    \multirow{2}{*}{\makecell{Threshold}} & \multicolumn{3}{c}{mAP}\\
    \cmidrule(rl){2-4}
    & @0.5 & @0.75 & \multicolumn{1}{c}{Avg.} \\
    % TaskWeave~\cite{yang2024task} & 56.51 & 33.66 \\
    \hline
    \addlinespace[2pt]
    0 & 66.99 & 47.89 & 46.85 \\
    0.5 & \textbf{67.87} & 47.01 & 46.50 \\
    0.6 & 67.79 & 47.40 & 46.74 \\
    0.7 & 67.64 & 47.97 & \textbf{46.98} \\
    \rowcolor{mygray}
    \textbf{0.8} & 67.38 & \textbf{48.00} & 46.95 \\
    0.9 & 67.17 & 47.96 & 46.92 \\
    \toprule
    \end{tabular}
\caption{Ablation study on the NMS threshold.}
\label{table9}
% \vspace{-0.2cm}
\end{center}
\end{table}

\begin{table}[t]
\begin{center}
\renewcommand{\arraystretch}{1.0}
\begin{tabular}{cccc}
    \toprule
    Loss type &  R1@0.5 & R1@0.7 & mAP$_{avg}$ \\
    % \noalign{\smallskip}
    \hline
    % \noalign{\smallskip}
    \addlinespace[2pt]
    Huber Loss & \textbf{68.00} & 51.94 & 46.68  \\
     L1 Loss & 66.65 & 51.89 & 46.73  \\
    \rowcolor{mygray}
     L2 Loss & 67.68 & \textbf{52.90} & \textbf{46.95} \\
    \addlinespace[1pt]
    \toprule
\end{tabular}
\caption{Ablation study on the IoU loss type.}
\label{table10}
\vspace{-0.2cm}
\end{center}
\end{table}

\subsubsection{Training Details}
\label{B2}

The cross-modal alignment encoder and region-guided decoder consist of three layers of transformer blocks. In the encoder, we also use a simple visual feature refinement module and a local regular loss following~\cite{sun2024tr}. The balancing parameters are set as: $\lambda_{\text{align}} = 0.3$, $\lambda_{\text{iou}} = 1$, $\lambda_{\text{L1}} = 10$, $\lambda_{\text{gIoU}} = 1$, $\lambda_{\text{cls}} = 4$, and $\lambda_{\text{sal}}$ is set as 1 for QVHighlights, 4 for Charades-STA and TACoS. The saliency margin $\Delta$ is set to 0.2, the temperature scaling parameter $\tau$ is set to 0.07, and the empirical hyperparameter $\alpha$ is set to 0.25 and $\gamma$ is set to 2. Non maximum suppression (NMS) with a threshold of 0.8 is applied during the inference for all three datasets. All experiments are implemented in Pytorch v1.13.0 on a single NVIDIA RTX 3090 GPU.

\subsection{Additional Experiments}
\label{C}

We compare our RGTR with the following state-of-the-art methods: 2D-TAN~\cite{zhang2020learning}, FVMR~\cite{gao2021fast}, MMN~\cite{wang2022negative}, SSRN~\cite{zhu2023rethinking}, Moment-DETR~\cite{lei2021detecting}, UMT~\cite{liu2022umt}, QD-DETR~\cite{moon2023query}, MomentDiff~\cite{li2024momentdiff}, UniVTG~\cite{lin2023univtg}, TR-DETR~\cite{sun2024tr}, TaskWeave~\cite{yang2024task}, UVCOM~\cite{xiao2024bridging}, CG-DETR~\cite{moon2023correlation}, LLMEPET~\cite{jiang2024prior}.

\begin{figure*}[ht]
    \centering
    \includegraphics[height=12.6cm]{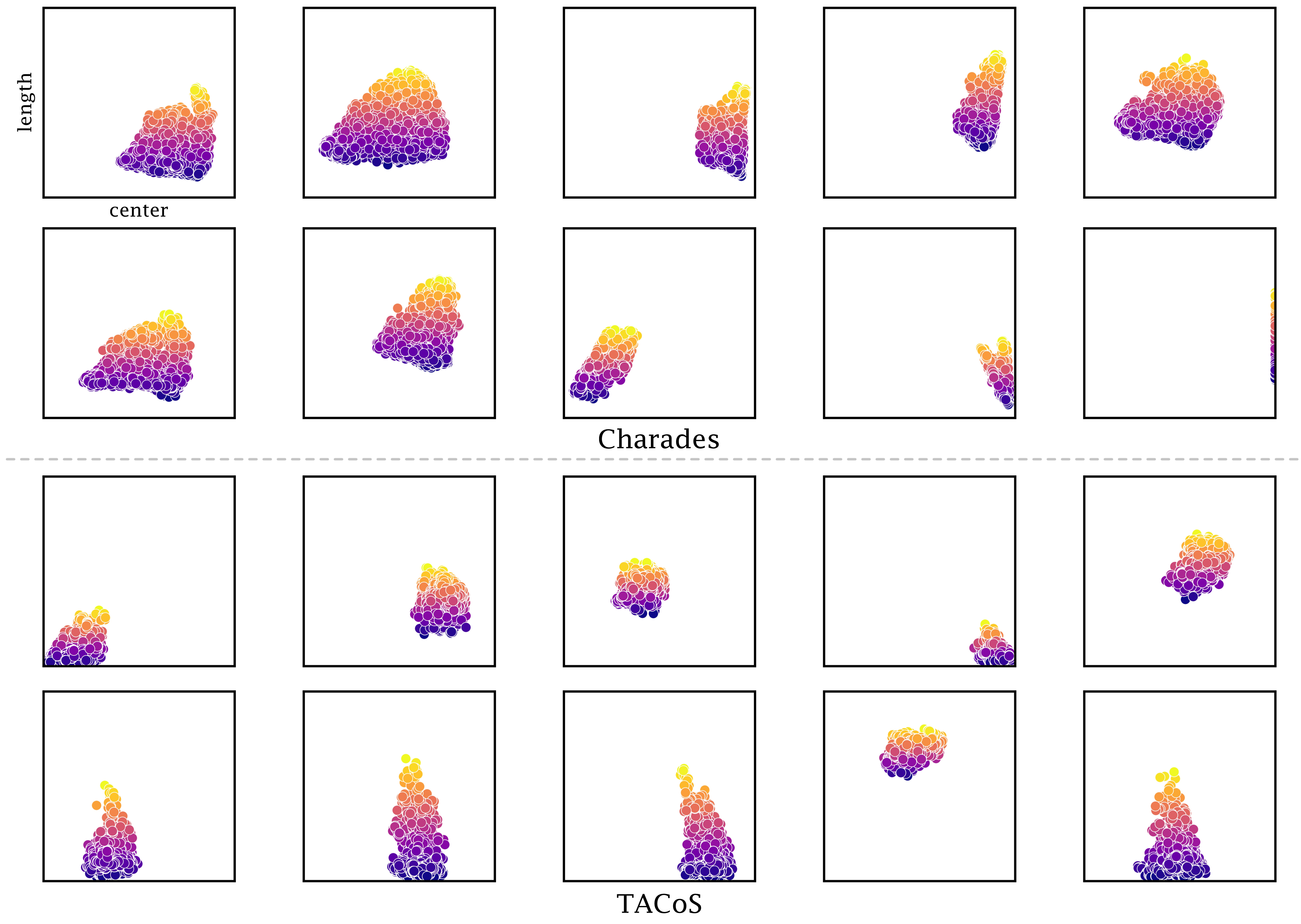}
    \caption{Visualization of all moment span predictions for all the videos on Charades and TACoS, for all the 10 moment queries in the region-guided decoder.}
    \label{fig:8}
\end{figure*}

\subsubsection{Experiments on Highlight Detection}
We also report the highlight detection results on QVHighlights ~\textit{test} and ~\textit{val} splits using mAP and HIT@1 in Tab.~\ref{table6}, although it is not the task of our interest. Our main contributions focus on the temporal grounding part (~\textit{i.e.}, the design of the decoder and moment queries), while highlight detection mainly depends on the learning of cross-modal interaction in the encoder part. As shown in Tab.~\ref{table6}, compared to other methods focusing on complex modal interaction, RGTR also maintains its competitiveness. The results indicate that regional guidance also indirectly helps highlight detection.

\subsubsection{Experiments on Charades-CD}
We conduct another out-of-distribution (OOD) experiment following~\cite{yuan2021closer}, which repartitions the Charades-STA dataset based on moment annotation density values. As shown in Tab.~\ref{table7}, similar to its performance on Charades-STA-Len and Charades-STA-Mom, RGTR also achieves the best performance on Charades-CD. These results demonstrate the robustness of RGTR in handling out-of-distribution scenarios.

\subsubsection{Experiments about VGG Features}
For Charades-STA, we also extract visual features with VGG~\cite{simonyan2014very} and use Glove~\cite{pennington2014glove} for textual features. As shown in Tab.~\ref{table8}, RGTR also achieves the best performance on Charades-STA using VGG features.

\subsubsection{Experiments about NMS Threshold}
As shown in Tab.~\ref{table9}, we perform ablation experiments on the NMS threshold. 
Because NMS does not affect the Rank-1 results, we only show the mAP results. Different NMS thresholds impact performance. We select the threshold as 0.8 based on mAP@0.75, prioritizing high IoU performance, although the performance at 0.7 threshold is also excellent. On the test set, performance at 0.8 threshold surpasses that at 0.7.

\subsubsection{Experiments about IoU Loss Types}
As shown in Tab.~\ref{table10}, we employ different IoU loss to supervise the IoU score. All loss types can significantly improve performance, among which L2 loss achieves optimal performance.

\subsection{Additional Visualization Results}
\label{D}

\begin{figure}[ht]
    \centering
    \includegraphics[height=8.4cm]{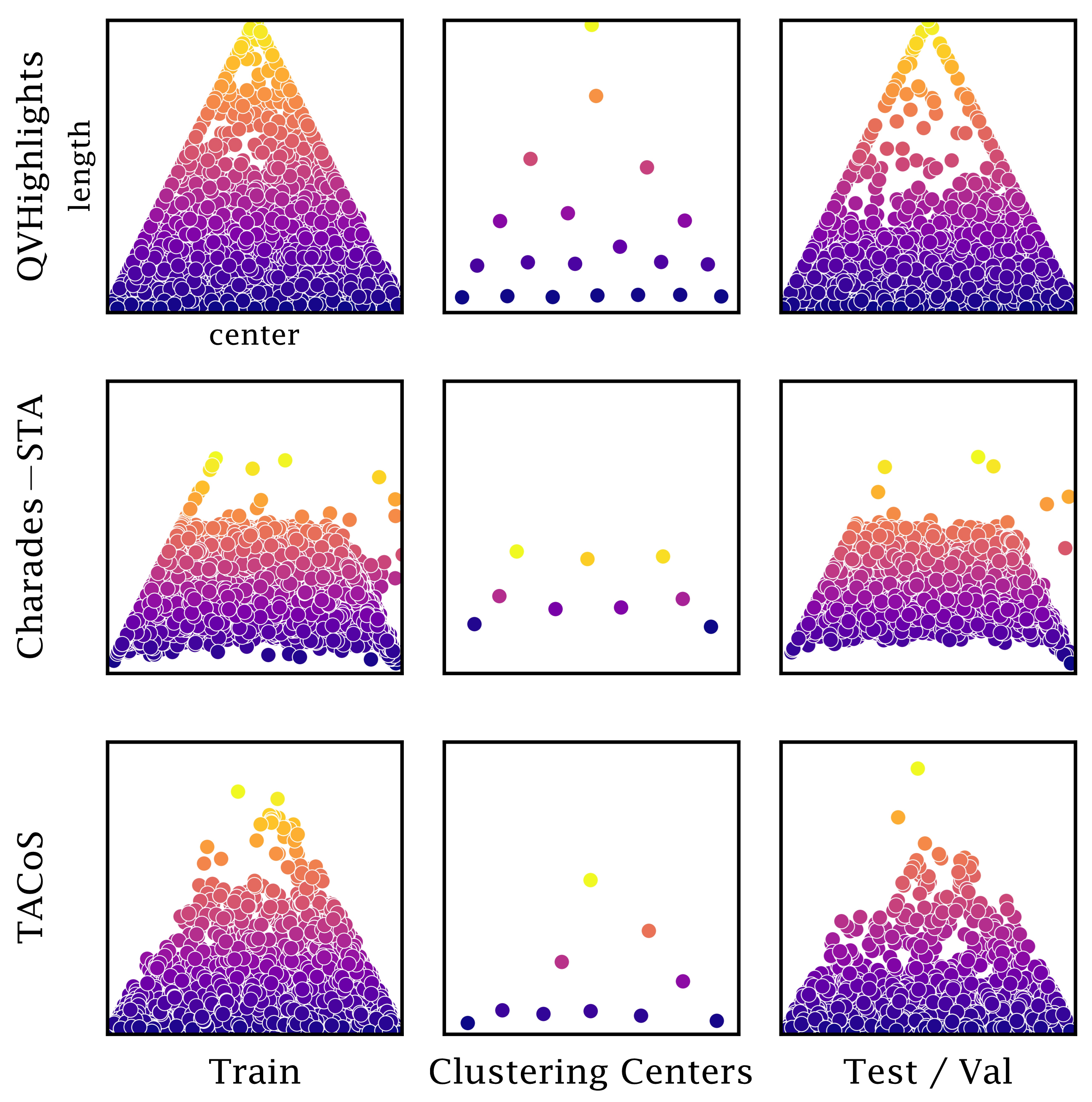}
    \caption{The distribution of training set, test set (validation set for QVHighlights) and clustering centers on QVHighlights, Charades-STA and TACoS.}
    \label{fig:9}
\end{figure}

\subsubsection{Visualization on Charades and TACoS}
We also present the visualization of moment span predictions on Charades and TACoS, for all the 10 dynamic anchors in the region-guided decoder. As shown in Fig.~\ref{fig:8}, due to the biased distribution on Charades-STA and TACoS, the diversity of moment queries is not obvious compared to QVHighlights, but we can still observe that they tend to cover as much and diverse temporal regions as possible.

\subsubsection{Distribution of Clustering Centers}
We present the distribution of clustering centers on three datasets in Fig.~\ref{fig:9}. The clustering centers generated by the training set distribution show great diversity and represent different temporal regions. These clustering centers represent a general distribution, approximating the center and length of most videos in life, allowing us to utilize them as regional priors for the test/val set. The diversity and representativeness of these clustering centers are the key to constructing regional guidance. In addition, as mentioned in Experiments, the biased distribution on Charades-STA and TACoS leads to less diversity learned by anchor pairs, resulting in minor performance improvement.

\subsection{Limitations}
\label{E}
In this paper, we utilize clustering centers from the distribution of all ground-truth moment spans as regional guidance. However, this regional guidance relies on manual clustering, and the quality of clustering significantly affects performance. This handcrafted technique, although effective, somewhat conflicts with the end-to-end philosophy of the DETR structure. Therefore, designing a kind of adaptive task-related guidance without handcrafted technique will be the focus of our future work.

\end{document}